%% file: eccv2022submission.tex
\documentclass[runningheads]{llncs}

\usepackage{graphicx}
\usepackage{tikz}
\usepackage{comment}
\usepackage{amsmath,amssymb}

\usepackage{color}
\usepackage[accsupp]{axessibility}  

\usepackage{cite}
\usepackage{xspace}
\usepackage{booktabs}
\usepackage{epsfig}
\usepackage{algorithm}
\usepackage{algpseudocode}
\usepackage{bm}
\usepackage{caption}
\usepackage{subcaption}
\usepackage[normalem]{ulem}
\usepackage{makecell}
\usepackage{multirow}
\usepackage{xcolor,colortbl}
\definecolor{Gray}{gray}{0.9}
\usepackage[pagebackref,breaklinks,colorlinks]{hyperref}
\usepackage[capitalize]{cleveref}
\crefname{section}{Sec.}{Secs.}
\Crefname{section}{Section}{Sections}
\Crefname{table}{Table}{Tables}
\crefname{table}{Tab.}{Tabs.}
\usepackage[misc]{ifsym}

\def\onedot{\ifx\@let@token.\else.\null\fi\xspace}
\def\eg{\emph{e.g}\onedot} 
\def\eg{\emph{e.g}\onedot} 
\def\ie{\emph{i.e}\onedot} 
 
\def\etc{\emph{etc}\onedot} \def\vs{\emph{vs}\onedot}
 
\def\etal{\emph{et al}\onedot}
\def\st{\emph{s.t}\onedot} 
\makeatother

\def\Vec#1{{\boldsymbol{#1}}}
\def\Mat#1{{\boldsymbol{#1}}}

\usepackage{pifont}
\DeclareMathOperator*{\argmin}{arg\,min}

\begin{document}
\pagestyle{headings}
\mainmatter
\def\ECCVSubNumber{4343}  

\title{Learning Instance and Task-Aware Dynamic Kernels for Few-Shot Learning} 



\titlerunning{INSTA}
%
%
\authorrunning{R. Ma et al.}

\author{Rongkai Ma\inst{1},  Pengfei Fang\inst{1,2,3}$\textsuperscript{(\Letter)}$,  Gil Avraham\inst{4},  Yan Zuo\inst{3}, \\ Tianyu Zhu\inst{1}, Tom Drummond\inst{5}, Mehrtash Harandi\inst{1,3}} 

\institute{\textsuperscript{\rm 1}Monash University
\;
\textsuperscript{\rm 2}Australian National University
\;
\textsuperscript{\rm 3}CSIRO
\\
\textsuperscript{\rm 4}Amazon Australia
\;
\textsuperscript{\rm 5}The University of Melbourne
\\
\Letter: Corresponding author
}

%

\maketitle

\input{Abstract/abstract}
\input{Intro/introduction}
\input{Related_work/related_work}
\input{Method/method_2}
\input{Experiments/exp}

\input{Conclusion/conclusion}
\input{Appendix/appendix}

%
%
\bibliographystyle{splncs04}
\bibliography{egbib}
\end{document}

%% file: Abstract/abstract.tex
\begin{abstract}
Learning and generalizing to novel concepts with few samples (Few-Shot Learning) is still an essential challenge to real-world applications. A principle way of achieving few-shot learning is to realize a model that can rapidly adapt to the context of a given task. Dynamic networks have been shown capable of learning content-adaptive parameters efficiently, making them suitable for few-shot learning. In this paper, we propose to learn the dynamic kernels of a convolution network as a function of the task at hand, enabling faster generalization. To this end, we obtain our dynamic kernels based on the entire task and each sample, and develop a mechanism further conditioning on each individual channel and position independently. This results in dynamic kernels that simultaneously attend to the global information whilst also considering minuscule details available. We empirically show that our model improves performance on few-shot classification and detection tasks, achieving a tangible improvement over several baseline models. This includes state-of-the-art results on four few-shot classification benchmarks: \emph{mini}-ImageNet, \emph{tiered}-ImageNet, CUB and FC100 and competitive results on a few-shot detection dataset: MS COCO-PASCAL-VOC.
\end{abstract}  

%% file: Intro/introduction.tex
\section{Introduction}
Despite the great success of the modern deep neural networks (DNNs), in many cases, the problem of adapting a DNN with only a handful of labeled data is still challenging. Few-Shot Learning (FSL) aims to address the inefficiencies of modern machine learning frameworks by adapting models trained on large databases for novel tasks with limited data~\cite{finn2017model, sung2018learning, nichol2018first, zhang2020deepemd, ma2022adaptive}.

Early approaches of FSL learned a fixed embedding function to encode samples into a latent space, where they could be categorized by their semantic relationships~\cite{snell2017prototypical, sung2018learning, zhang2020deepemd}. However, such fixed approaches do not account for category differences~\cite{ren2018meta}, which may exist between already learned tasks and novel tasks. Ignoring these discrepancies can severely limit the adaptability of a model as well as its ability to scale in the FSL setting. Although methods that adapt embeddings~\cite{vinyals2016matching, lu2021tailoring, ye2020few} attempt to address this issue, they still utilize fixed models which lack full adaptability and are constrained to previously learned tasks. Another group of approaches, such as~\cite{finn2017model, nichol2018first}, adapt models with a few optimization steps. Given the complexity of the loss landscape of a DNN, such methods come short compared to metric-based solutions~\cite{wang2020generalizing}.

\begin{figure}[t!]
\centering
\includegraphics[width=0.95\columnwidth]{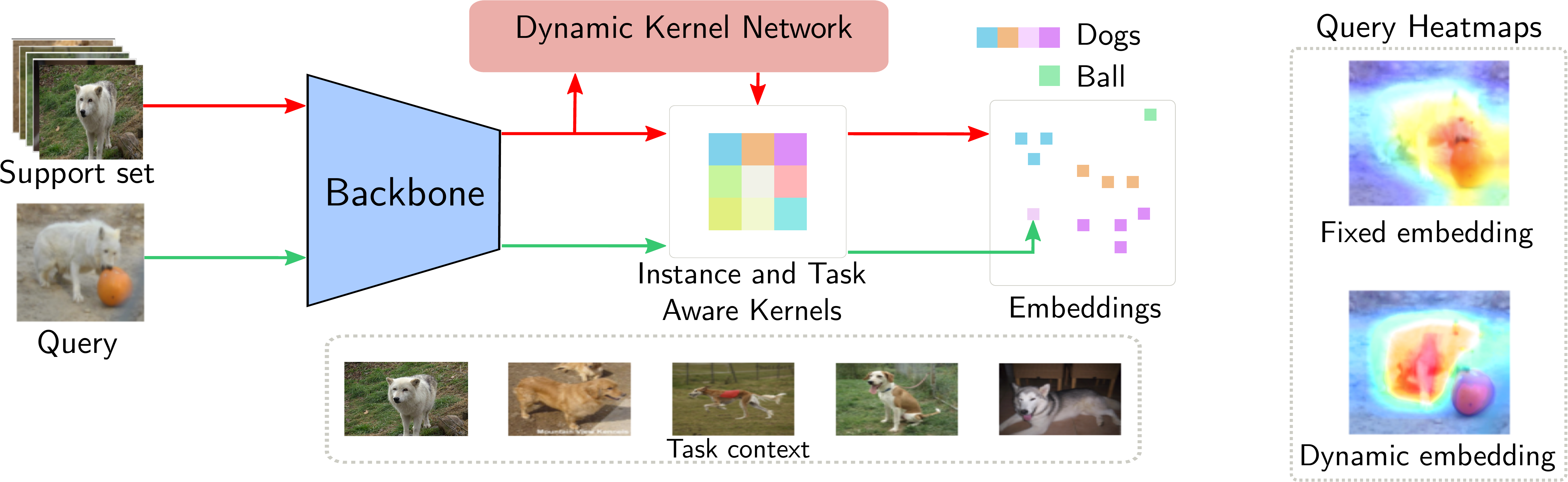}
\caption{Typical FSL models learn fixed embeddings, which are not flexible enough to rapidly adapt to novel tasks. Our method instead uses a dynamic kernel network to produce a set of dynamic parameters which are both instance and task-aware
}
\label{fig:conceptual}
\end{figure}

Dynamic kernel approaches have been shown to be computationally efficient~\cite{zhou2021decoupled} relative to optimization-based solutions, resulting in the models capable of encoding the novel information at the parameter level~\cite{jia2016dynamic}. A very recent study~\cite{xu2021learning} showed that dynamic kernel methods are effective when applied to FSL via the learning of input-conditioned filters, enabling the realization of adaptive models; a key limitation of this dynamic kernel method is that only the per-class level information is utilized via class prototypes.

Arguably, when learning new tasks with sparse data, it is vital to fully utilize information on both an instance and task-level for efficiency. As an example, when given the task of differentiating between dog breeds (Fig.~\ref{fig:conceptual}), both task and instance-level information is required; it is important that we are sensitive enough to distinguish the minuscule details between different dog breeds (instance-level), yet can still focus on the global knowledge required to filter out irrelevant, non-task related objects (task-level).

In this paper, we propose a novel, dynamic approach for FSL tasks. Our method is realized by a dynamic kernel that can encode both instance-level and task-level features via jointly learning a dynamic kernel generator and a context learning module. The dynamic kernel generator adaptively produces an instance kernel by exploiting the information along the spatial and channel dimensions of a feature map in a decoupled manner. It further incorporates information from the frequency domain resulting in a rich set of descriptive representations. The context learning module refines the support features into a task-specific representation which is used to produce the task-specific kernel. The resulting instance and task-specific kernels are fused to obtain a dynamic kernel which is \underline{INS}tance and \underline{T}ask-\underline{A}ware (INSTA). Our method differentiates from approaches such as FEAT~\cite{ye2020few} and GLoFa~\cite{lu2021tailoring} by learning a set of dynamic parameters adaptive to the novel tasks instead of employing fixed models during inference. Furthermore, in contrast to optimization-based methods~\cite{finn2017model}, our approach can adapt the model parameters without the requirements of backpropagation during the inference stage. We offer the following contributions:
\begin{itemize}
    \item We propose a novel FSL approach to extract both instance and task-specific information using dynamic kernels.
     
    \item We offer the first FSL framework capable of being evaluated on both classification \emph{and} detection tasks.

    \item Empirically, we offer substantial improvements over several FSL baselines, including optimization-based and metric-based approaches.
\end{itemize}

%% file: Related_work/related_work.tex
\section{Related Work}\label{sec:related_work}
The family of few-shot learning literature is broad and diverse. However, those related to this work are mainly the family of optimization-based methods~\cite{finn2017model,flennerhag2019meta,franceschi2018bilevel,nichol2018first,rusu2018meta,lee2019meta, simon2020modulating,bertinetto2018meta,9533455} and metric-based methods~\cite{bertinetto2016learning,simon2020adaptive,snell2017prototypical,sung2018learning,vinyals2016matching,ye2020few,zhang2020deepemd,fang2021kernel}.

\noindent{\bf{Optimization based methods.}} Optimization-based methods such as MAML~\cite{finn2017model} or Reptile~\cite{nichol2018first} focus on learning a set of initial model parameters that can generalize to new tasks (or environments) with a small number of optimization steps and samples without severe over-fitting. 
Furthermore, in most cases, this group of methods present a framework trained with a bi-level optimization setting\cite{franceschi2018bilevel}, which provides a feasible solution to adapt the model to the test set from the initialized model.

Our proposed framework is similar to the optimization-based
methods~\cite{antoniou2019train,andrychowicz2016learning,chen2019closer,lee2018gradient,liu2020ensemble} in the sense that the model parameters are task-adaptive. However, our solution does not require backpropagation during inference to achieve so. Furthermore, our method can be incorporated with optimization-based methods, and empirically we observed that such construction yields performance improvement. This will be demonstrated in \S~\ref{sec:exp}.

\noindent{\bf{Metric-based methods.}} 
In few-shot learning literature, the metric-based methods aim to define a metric to measure the dis/similarity between samples from a few observations~\cite{liu2018learning,satorras2018few,koch2015siamese,choi2018structured,xu2020attentional,zhao2021looking,liu2021learning,wertheimer2021few}. ProtoNet~\cite{snell2017prototypical} achieves this by learning a fixed latent space where the class representations (\ie, prototype), obtained by averaging the features from the same class, are distinctive for different classes. DeepEMD~\cite{zhang2020deepemd} formulates the query-support matching as an optimal transport problem and adopts Earth Mover's distance as the metric. One commonality in the aforementioned methods is the fact that all employ a fixed embedding space when facing novel tasks, which essentially limits their adaptability. On the other hand,
many previous methods suggested to adapt the embeddings to the novel tasks~\cite{shyam2017attentive, triantafillou2017few, ye2020few, lu2021tailoring, oreshkin2018tadam,hou2019cross,fei2020melr}.
CTM~\cite{li2019finding} proposes to produce a mask to disregard the uninformative features from the support and query embeddings during inference. MatchingNet~\cite{vinyals2016matching} uses a memory module to perform sample-wise embedding adaptation and determines the query label by a cosine  distance.
TADAM~\cite{oreshkin2018tadam} proposes to learn a dynamic feature extractor by applying a linear transformation to the embeddings to learn a set of scale and shift parameters. FEAT~\cite{ye2020few} and GLoFa~\cite{lu2021tailoring} provide an inspiring way to perform the embedding adaptation using a set function. 

Our contribution is complementary to the aforementioned methods. We aim to learn a set of dynamic kernels via exploiting the instance and task-level information according to the task at hand. This results in a more distinctive and descriptive representation, which effectively boosts the performance of these methods directly relying on constructing a metric space.

\noindent{\bf{Dynamic kernels.}}
The application of dynamic kernels solutions within the domain of few-shot learning
is less explored in the current literature. However, it has been demonstrated useful when labels are abundant~\cite{chen2020dynamic,zhou2021decoupled, ha2016hypernetworks,huang2018multi, teerapittayanon2016branchynet, bolukbasi2017adaptive,wang2018skipnet,veit2018convolutional}. Zhou~\etal~\cite{zhou2021decoupled} have proposed to use decoupled attention over the channel and spatial dimensions. This results in a content-adaptive and efficient method that provides a feasible way to achieve task-adaptive FSL.

To leverage the effectiveness of the dynamic kernel into few-shot learning tasks, \cite{xu2021learning} propose to learn dynamic filters for the spatial and channels separately via grouping strategy. The resulted kernels are then applied to the query to produce a support-aligned feature. Due to the usage of grouping, the performance might sacrifice for efficiency~\cite{zhou2021decoupled}. Inspired by these methods, our INSTA produces dynamic kernels that are both instance-aware and task (or episodic)-aware while also incorporating valuable frequency patterns; as such, our method produces more informative and representative features. 

%% file: Method/method_2.tex
\section{Method}
In this section, we introduce our proposed INSTA. Note that while we present our method in terms of few-shot classification, the proposed approach is generic and can be seamlessly used to address other few-shot problems, including structured-prediction tasks such as few-shot object detection (see \S~\ref{ssec:fsd} for details). 

\subsection{Problem Formulation}
\begin{figure*}[t] 
\begin{center}
\includegraphics[width=\linewidth]{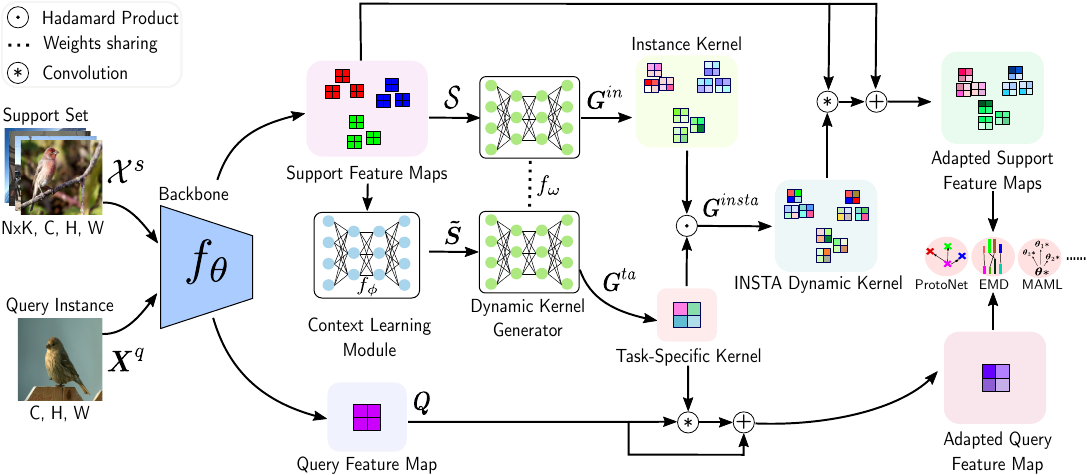}
\end{center}
\caption{The framework of our method. Given the support set $\mathcal{X}^s$ and the query sample $\Mat{X}^q$, the backbone network $f_{\theta}$ first encodes them into a representation space as $\mathcal{S}=\{\Mat{S}_{11},\ldots,\Mat{S}_{NK}\}$ and $\Mat{Q}$. Then a dynamic generator $f_{\omega}$ is used to produce an instance kernel for each support sample. Meanwhile, a context learning module $f_{\phi}$ is used to refine and aggregate the context of the entire support feature maps to produce a task-specific representation, which is then used as the input of $f_{\omega}$ to obtain the task-specific kernel.
Note the instance kernel and task-specific kernel share the parameters across $f_{\omega}$.
Finally, the INSTA dynamic kernel that is both instance-aware and task-aware is applied to the support feature maps, while only the task-specific kernel is applied to the query feature map to obtain the adapted representations
} 
\label{fig:framework}
\end{figure*}

In what follows, we will give a brief description of the problem formulation for few-shot classification. The vectors and matrices (or high-dimensional tensors) are denoted by bold lower-case letters (\eg, $\Vec{x}$) and bold upper-case letters (\eg, $\Mat{X}$) throughout this paper. FSL aims to generalize the knowledge acquired by  a model from  a set of examples $\mathcal{D}_{train}=\{(\Mat{X}_i, y_i) | y_i\in\mathcal{C}_{train}\}$, to novel and unseen tasks $\mathcal{D}_{test}=\{(\Mat{X}_i, y_i) | y_i\in\mathcal{C}_{test}\}, \mathcal{C}_{train}\cap\mathcal{C}_{test}=\emptyset$, in a low data regime. 

We follow the meta-learning protocol to formulate FSL with episodic training and testing. Specifically, an episode $\mathcal{E}$ consists of a support set
$\mathcal{X}^s = \{ (\Vec{X}^s_{ij}, y^s_{i})| i = 1, \ldots, N, j = 1, \ldots, K, y_i^s\in\mathcal{C}_{train}\}$, where $\Vec{X}^s_{ij}$ denotes the $j$-th sample in the class $y^s_i$ and the query set $\mathcal{X}^q = \{(\Vec{X}^q_{i}, y^q_i) |  i = 1, \ldots, N\}$, where $\Vec{X}^q_{i}$ denotes a query example \footnote{Without losing generality, we use one sample per class as a query for presenting our method. In practice, each episode contains multiple samples per query class.} sampled from class $y^q_i$ (the test setup is the same as training but the episodes are sampled from $\mathcal{D}_{test}$). Such a formulation is 
known as $N$-way $K$-shot, where the goal is to utilize the support samples and their labels to obtain $\Theta^\ast$, the optimal parameters of the model, such that each query is classified correctly into one of classes in the support set. The object of network training follows that:
\begin{equation}\label{eq:objective}
\Theta^\ast = \argmin_{\Theta}\sum_{\Mat{X}^q_i\in\mathcal{X}^q}\mathcal{L}(f_{\Theta}(\Mat{X}^q_i|\mathcal{X}^s),y^q_i),
\end{equation}
where $f_{\Theta}$ represents the entire model parameterized by $\Theta$. 

\subsection{Model Overview}
We first provide an overview of our model (see the sketch of the pipeline in Fig.~\ref{fig:framework}). Overall, our framework is composed of three modules, the backbone network $f_{\theta}$, the dynamic kernel generator $f_{\omega}$, and the context learning module $f_{\phi}$. The backbone network first extracts the feature maps from the input images, followed by a two-branch dynamic kernel network to obtain our proposed dynamic kernels. Specifically, the dynamic kernel generator independently refines the features in the support set to produce the instance kernel (\ie, $\Mat{G}^{in}$) in one branch. In another branch, the context learning module $f_{\phi}$ first produces a task-specific representation by refining and summarizing the features from the entire support set, which is then used as the input of the dynamic kernel generator to produce the task-specific kernel (\ie, $\Mat{G}^{ta}$) adaptively. The dynamic kernel generator is shared across these two branches to generate the instance kernel and the task-specific kernel. Such a two-branch design enables the network to be aware of both the instance-level feature and global context feature of the support set. The design choice is justified in \S ~\ref{sec:exp}. Finally, both the instance and task-specific kernels are fused and employed to boost the discrimination of instance features in the support set. The task-specific kernel is applied to the query feature maps for refining the representation without extracting a query instance kernel. Given the adapted support and query features, any post-matching algorithm (\eg, metric-based or optimization-based FSL) can be employed seamlessly to achieve the few-shot classification task.

\subsection{Dynamic Kernel Generator}
\label{ssec:dynamic_kernel generator}
\begin{figure}[!ht]
\begin{center}
   \scalebox{1.1}{
   \includegraphics[width=0.9\linewidth]{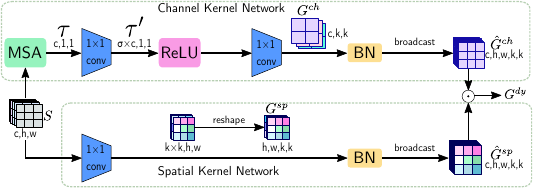}
}
\end{center}
   \caption{The architecture of dynamic kernel generator. We adopt a decomposed architecture to produce dynamic kernels for the channels and spatial dimensions independently. This results in a light-weight kernel, which greatly fits the low-data regime of FSL
   } 
\label{fig:dynamic_kernel_generator}
\end{figure}

In this part, we provide a detailed description of the dynamic kernel generator. The central component of our model is the dynamic kernel generator, which receives a feature map as input and produces a kernel adaptively. An essential problem one may face is that the feature map extracted by modern DNNs usually has a large size (\eg, $640\times 5\times 5$ for ResNet-12). As such, we need to identify $c_{out}\times c\times k\times k$ amount of parameters for the dynamic kernel, where the $c_{out}$ is the output channel size (we consider $c=c_{out}$ in our paper), $c$ is the input channel size, and $k$ is the kernel size. This poses a significant issue since the generated dynamic kernel is prone to overfitting given the limited data of FSL. To tackle this problem, we consider designing our dynamic kernel generator in a decomposed manner~\cite{zhou2021decoupled}. As such, we develop a channel kernel network and a spatial kernel network to produce a kernel for each channel and each spatial location independently. This design exploits the information per data-sample to a large extent while reducing the number of parameters of a dynamic kernel, which greatly fits the low data regime of FSL. Fig.~\ref{fig:dynamic_kernel_generator} illustrates the pipeline of the proposed dynamic kernel generator. In what follows, we detail out the operations of the channel and spatial kernel networks. 

\noindent{\bf{Channel Kernel Network.}} \label{para:channel_branch} 
Given a feature map $\Mat{S}\in\mathbb{R}^{c\times h\times w}$, where $c$, $h$, and $w$ denote the size of channel, height and width, a common practice to realize the channel kernel network follows the SE block~\cite{hu2018squeeze}, as done for example in~\cite{chen2020dynamic}. In the SE block,  global average pooling (GAP)  is performed over the feature map to encode the global representation, which is fed to the following sub-network for channel-wise scaling.
The drawback of this design is that the GAP operator mainly preserves the low frequency components of the feature map, as shown in~\cite{qin2020fcanet} (averaging is equivalent to low-pass filtering in the frequency domain). Thus, GAP discards important signal patterns in the feature map to a great degree.
Clearly, low-frequency components cannot fully characterize the information encoded in a feature map, especially in the low data regime. We will empirically show this in \S~\ref{subsec:ablation}. To mitigate this issue, we opt for  multi-spectral attention (MSA) to make better use of high-frequency patterns in the feature map. Given a feature map $\Mat{S}$,
we equally split the feature map into $n$ smaller tensors along the channel dimension (in our experiments  $n=16$), as $\{\Mat{S}^{0}, \Mat{S}^{1},\ldots,\Mat{S}^{n-1}|\Mat{S}^i\in\mathbb{R}^{\frac{c}{n}\times h\times w}\}$. Then, each channel of the  tensor $\Mat{S}^i$ is processed by a basis function of a 2D-\underline{D}iscrete \underline{C}osine \underline{T}ransform (DCT) following the work of Qin~\etal~\cite{qin2020fcanet}. As a result, we obtain a real-valued feature vector in the frequency domain as $\Vec{\tau}^i = \mathrm{DCT}(\Mat{S}^i),  \Vec{\tau}^i \in \mathbb{R}^{{c}/{n}}$, which is the frequency-encoded vector corresponding to $\Mat{S}^i$. The frequency-encode vector for $\Mat{S}$ is obtained by concatenating $\Vec{\tau}^i$s as:
\begin{equation}\label{eq:cat}
 \Mat{\tau} = \mathrm{concat}(\Mat{\tau}^0,\Mat{\tau}^1,\ldots,\Mat{\tau}^{n-1}),
\end{equation}
where $\Mat{\tau} \in\mathbb{R}^{c}$. 
Please refer to the supplementary material for the theoretical aspects of the MSA module. As compared to the global feature obtained by GAP, frequency-encoded feature $\Vec{\tau}$ contains more diverse information patterns, which brings extra discriminative power to our model. This will be empirically discussed in \S~\ref{subsec:ablation}.

Once we obtain $\Vec{\tau}$, a light-weight network is used to produce the channel kernel values adaptively. This network is realized by a two-layer MLP (or $1\times 1$ convolution), with architecture as $c\to \sigma\times c \to \mathrm{ReLu} \to k^2 \times c$, where $0<\sigma<1$, and $k$ controls the size of receptive field in the dynamic kernel (practically, we set $\sigma=0.2$ and $k=3$). Then we reshape the output vector into a $c\times k\times k$-sized tensor, denoted by $\Mat{G}^{ch}$. We can obtain the final channel dynamic kernel $\hat{\Mat{G}}^{ch} \in\mathbb{R}^{c\times h\times w\times k\times k}$ via applying batch normalization (BN) and spatial-wise broadcast to $\Mat{G}^{ch}$.

\noindent{\bf{Spatial Kernel Network.}}
Independent to the channel kernel network, our spatial kernel network adaptively produces a convolution kernel $\Mat{G}^{sp}_{a,b}\in\mathbb{R}^{k\times k}$ for each spatial location of a feature map (\ie, $\Mat{S}_{:,a,b}$). This is achieved by using a $1\times 1$ convolution layer, with the architecture of $(c, k^2, 1, 1)$ and the reshape operation. Therefore, the spatial kernel $\Mat{G}^{sp}$ for all the spatial locations of a feature map $\Mat{S}$ has a size of $h\times w\times k\times k$. We then obtain the final spatial dynamic kernel $\hat{\Mat{G}}^{sp}\in\mathbb{R}^{c\times h\times w\times k\times k}$ by applying the BN and channel-wise broadcast to $\Mat{G}^{sp}$.

Finally, to unify the channel and spatial dynamic kernels, we apply the Hadamard product between the $\hat{\Mat{G}}^{ch}$ and $\hat{\Mat{G}}^{sp}$ to obtain the dynamic kernel $\Mat{G}^{dy}$ as:
\begin{equation}\label{eq:instance_kernel}
    \Mat{G}^{dy} = \hat{\Mat{G}}^{sp}\odot\hat{\Mat{G}}^{ch},
\end{equation}
where $\Mat{G}^{dy}\in\mathbb{R}^{c\times h\times w\times k\times k}$. Notably, the number of parameters of our dynamic kernel is much less than a normal convolution kernel ($c\times h\times w\times k^2 \ll c_{out}\times c\times k^2$ given $c_{out}=c=640$ and $h=w=5$).

\subsection{Dynamic Kernel}
In this part, we discuss the process of obtaining the proposed INSTA dynamic kernel, which consists of the instance kernel and the task-specific kernel. 

\noindent{\textbf{Instance Kernel.}}\label{sssec:instance kernel} 
We defined the instance kernel as the dynamic kernel extracted from each support feature map. Formally, given the support set images $\mathcal{X}^s=\{\Mat{X}^s_{11},\ldots,\Mat{X}^s_{NK}|\Mat{X}^s_{ij}\in\mathbb{R}^{C\times H\times W }\}$, where $C$, $H$, and $W$ indicate the channel, height, and width of an image, respectively, the backbone network first extracts a feature map from each image, as $\mathcal{S}=f_{\theta}(\mathcal{X}^s)$, with $\mathcal{S}=\{\Mat{S}_{11},\ldots,\Mat{S}_{NK}|\Mat{S}_{ij}\in\mathbb{R}^{c\times h\times w}\}$. The feature map for the query sample can be obtained in a similar fashion, as $\Mat{Q}_i = f_{\theta}(\Mat{X}_i^q)$. We then use the dynamic kernel generator $f_{\omega}$ to independently produce a dynamic kernel $\Mat{G}^{in}_{ij}\in\mathbb{R}^{c\times h\times w\times k\times k}$ for each support feature map $\Mat{S}_{ij}$, following the process described in \S~\ref{ssec:dynamic_kernel generator}. 

\noindent{\textbf{Task-Specific Kernel.}}
Adapting according to the whole context of support set is essential for FSL since it contains essential information of the task~\cite{oreshkin2018tadam}.
Following this intuition, we propose to learn the task-specific kernel to represent the knowledge of the task encoded in the support set. We achieve this by first using a context learning module $f_{\phi}$ to produce a fully context-aware representation for the support set by refining and aggregating the intermediate features. To be specific, we use four $1\times1$ convolution layers with a summation layer in the middle of the network to aggregate the $N\times K$ support features into one task-specific representation $\Tilde{\Mat{S}}$ (Please refer to the supplementary material for the conceptual diagram of $f_{\phi}$). 
Formally, this task representation can be obtained by:
\begin{equation}\label{eq:task_adaptation}
\Tilde{\Mat{S}} = f_{\phi}(\mathcal{S}),    
\end{equation}
where $\Tilde{\Mat{S}}\in\mathbb{R}^{c\times h\times w}$ and the $\mathcal{S}$ denotes the entire support set. Then this task-specific representation $\Tilde{\Mat{S}}$ is used as the input to the dynamic kernel generator, following the process described in \S~\ref{ssec:dynamic_kernel generator} to adaptively produce our task-specific kernel $\Mat{G}^{ta}\in\mathbb{R}^{c\times h\times w\times k\times k}$.

\subsubsection{INSTA Dynamic Kernel.}
\label{sssec:dynamic kernel}
Once we have the instance and task-specific kernels, we fuse each instance kernel $\Mat{G}_{ij}^{in}$ 
with task-specific kernel $\Mat{G}^{ta}$ using the Hadamard product, such that each dynamic kernel considers both the instance-level and task-level features. Formally, it can be described as: 
\begin{equation}\label{eq:dynamic_kernel}
\Mat{G}_{ij}^{insta} = \Mat{G}_{ij}^{in}\odot\Mat{G}^{ta},
\end{equation}
where $\Mat{G}_{ij}^{insta}$ is the final fused kernel corresponding to the $ij$-th sample in the support set.

After $\Mat{G}_{ij}^{insta}$ is obtained, we apply it on its corresponding support feature map $\Mat{S}_{ij}$. This is equivalent to first apply the task-specific kernel to extract the features, which is relevant to the task and then apply the instance kernel to further increase the discriminatory power of the resulting feature maps. While we only apply the task-specific kernel to the query feature map. This design choice is justified in \S~\ref{subsec:ablation}, where we compare our current design with a variant where we also extract instance kernel from the query feature maps.
The dynamic convolution can be implemented using the Hadamard product between the unfolded feature map and the dynamic kernel.
In doing so, the unfold operation first samples a $k\times k$ spatial region of the input feature map at each time and then stores the sampled region into extended dimensions, thereby obtaining an unfolded support feature map $\Mat{S}_{ij}^u\in\mathbb{R}^{c\times h\times w\times k\times k}$ and an unfolded query feature map $\Mat{Q}_{i}^u\in\mathbb{R}^{c\times h\times w\times k\times k}$. Then the adapted support and query feature maps are obtained as:
\begin{equation}\label{eq:conv}
\begin{split}
\dot{\Mat{S}}_{ij} = \text{AvgPool2d}(\Mat{S}^{u}_{ij}\odot\Mat{G}^{insta}_{ij}) + \Mat{S}_{ij},
\\
\dot{\Mat{Q}}_{i} = \text{AvgPool2d}(\Mat{Q}^{u}_{i}\odot\Mat{G}^{ta})+ \Mat{Q}_i,
\end{split}
\end{equation}
where $\dot{\Mat{S}}_{ij}\in\mathbb{R}^{c\times h\times w}$ and $\dot{\Mat{Q}}_{i}\in\mathbb{R}^{c\times h\times w}$. Note that the convolution operation between the dynamic kernel and the feature map in Fig.~\ref{fig:framework} is equivalent to the average of the Hadamard product between the unfolded feature map and the dynamic kernel over the $k\times k$ dimension (see Fig.~\ref{fig:diff}). In our design, we adopt a residual connection between the original and the updated feature maps to obtain the final adapted features. Having the adapted support and query feature maps, any post-matching algorithms can be adopted to achieve FSL tasks. In ~\S~\ref{sec:exp}, we demonstrate that our algorithm can further boost the performance on various few-shot classification models (\eg, MAML~\cite{finn2017model}, ProtoNet~\cite{snell2017prototypical} and EMD~\cite{zhang2020deepemd}) and the few-shot detection model \cite{fan2020few}.

\begin{remark}
The proposed dynamic kernels are convolutional filters and hence by nature differ from attention masks. As shown in Fig.~\ref{fig:diff}, the attention mask merely re-weights each element of a feature map~\cite{AIA_Pengfei_PAMI, hu2018squeeze}. In contrast, our proposal in Eq.~\eqref{eq:conv} realizes the dynamic convolution by first performing the element-wise multiplication between the unfolded feature map and the dynamic kernels, whose results are then averaged over the unfolded dimension. This is essentially equivalent to convolution operation (see Fig.~\ref{fig:diff}).




\begin{figure}[!ht]
\begin{center}
  \scalebox{1}{
  \includegraphics[width=0.9\linewidth]{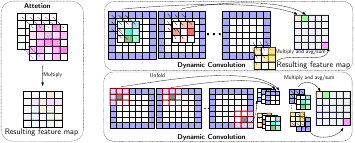}
}
\end{center}
  \caption{Schematic comparison of the dynamic convolution and attention mechanism. The 2-D average of the element-wise multiplication between unfolded features and dynamic kernels is equivalent to using adaptive kernels sliding over the original feature map} 
\label{fig:diff}
\end{figure}

\end{remark}


%% file: Experiments/exp.tex
\section{Experiments}\label{sec:exp}
In this section, we first evaluate our method across four standard few-shot classification benchmarks: \emph{mini}-ImageNet~\cite{Ravi2017OptimizationAA}, \emph{tiered}-ImageNet~\cite{ren2018meta}, CUB~\cite{wah2011caltech}, and FC100~\cite{oreshkin2018tadam}. Full details of the implementation are provided in the supplementary material. Furthermore, we evaluate the effectiveness of our framework for the few-shot detection task on the MS COCO and PASCAL VOC  datasets~\cite{fan2020few}. Finally, we provide an ablation study to discuss the effect of each module in our framework. Please refer to supplementary material for a detailed description of each dataset.

\subsection{Few-Shot Classification}
We conduct few-shot classification experiments on three different state-of-the-art models, including MAML~\cite{finn2017model}, ProtoNet~\cite{snell2017prototypical} and DeepEMD~\cite{zhang2020deepemd} as baselines, and employ the proposed INSTA on top of them.
We use the ResNet-10 backbone for the MAML and the ResNet-12 backbone for the other two baselines across all four benchmarks. 
For a fair comparison, we implement all the baseline models to report the results indicated by ``*" across Table~\ref{tab:mini-tiered}~--~Table~\ref{tab:CUB-FC100}. Notably, for DeepEMD experiments, we adopt the \emph{open-cv} solver instead of the \emph{qpth} solver originally used in the paper to train the network due to resource capacity, which is indicated by ``$\clubsuit$". Moreover, following the the same evaluation protocols in our baseline frameworks~\cite{chen2019closer, zhang2020deepemd}, we report the mean accuracy with $95\%$ confidence interval (Please refer to supplementary material for more implementation details).

\begin{table}[!h]
    \caption{Few-shot classification accuracy and 95\% confidence interval on \emph{mini}-ImageNet and \emph{tiered}-ImageNet with ResNet backbones 
    }
    \begin{center}
    \scalebox{0.89}{
    \begin{tabular}{l c| c c|c c}
        \Xhline{2\arrayrulewidth}
         \multirow{2}{*}{Model}                      
         &\multirow{2}{*}{Backbone}
         &\multicolumn{2}{c|}{\emph{mini}-ImageNet}
         &\multicolumn{2}{c}{\emph{tiered}-ImageNet}
         \\
         
         &
         
         &{5-way 1-shot}              &{5-way 5-shot}  
         &{5-way 1-shot}              &{5-way 5-shot}    
        \\
        \Xhline{2\arrayrulewidth}

        MetaOptNet~\cite{lee2019meta}  
        &ResNet-12
        &$62.64\pm{0.61}$                      
        &$78.63\pm{0.46}$
        &$68.23\pm{0.23}$                   
        &$84.03\pm{0.56}$
        \\
        
        TADAM~\cite{oreshkin2018tadam}
        &ResNet-12
        &$58.50\pm{0.30}$                      
        &$76.70\pm{0.30}$
        &-
        &-
        \\
       
        FEAT~\cite{ye2020few}                        
        &ResNet-12
        &$66.78\pm{0.20}$                      
        &$82.05\pm{0.14}$
        &$70.80\pm{0.23}$                      
        &$84.79\pm{0.16}$
        
        \\
        CAN~\cite{hou2019cross}                       
        &ResNet-12
        &$63.85\pm{0.48}$         
        &$79.44\pm{0.34}$
        &$69.89\pm{0.51}$                      
        &$84.23\pm{0.37}$
        \\
    
        FRN~\cite{wertheimer2021few}                      
        &ResNet-12
        &$66.45\pm{0.19}$       
        &$82.83\pm{0.13}$
        &$72.06\pm{0.22}$                      
        &$86.89\pm{0.14}$
        \\

        InfoPatch~\cite{liu2021learning}                
        &ResNet-12
        &$67.67\pm{0.45}$       
        &$82.44\pm{0.31}$
        &$71.51\pm{0.52}$       
        &$85.44\pm{0.35}$
        \\
        GLoFA~\cite{lu2021tailoring}                   
        &ResNet-12
        &$66.12\pm{0.42}$       
        &$81.37\pm{0.33}$
        &$69.75\pm{0.33}$                      
        &$83.58\pm{0.42}$
        \\
       
        DMF~\cite{xu2021learning}                      
        &ResNet-12
        &$67.76\pm{0.46}$       
        &$82.71\pm{0.31}$
        &$71.89\pm{0.52}$                      
        &$85.96\pm{0.35}$
        \\
        
        \Xhline{2\arrayrulewidth}
        
        MAML*~\cite{finn2017model} 
        &ResNet-10
        &$54.73\pm{0.87}$
        &$66.72\pm{0.81}$ 
        &$59.85\pm{0.97}$                   
        &$73.20\pm{0.81}$
        \\
        
        \rowcolor{Gray}
        \textbf{INSTA-MAML*}  
        &ResNet-10
        &$\bm{56.41\pm{0.87}}$
        &$\bm{71.56\pm{0.75}}$  
        &$\bm{63.34\pm{0.92}}$                    
        &$\bm{78.01\pm{0.71}}$
        \\
        \Xhline{2\arrayrulewidth}
        ProtoNet*~\cite{snell2017prototypical}                    
        &ResNet-12
        &$62.29\pm{0.33}$            
        &$79.46\pm{0.48}$ 
        &$68.25\pm{0.23}$                   
        &$84.01\pm{0.56}$
      
        \\ 
        \rowcolor{Gray}
        \textbf{INSTA-ProtoNet*}              
        &ResNet-12
        &$\bm{67.01\pm{0.30}}$                      &$\bm{83.13\pm{0.56}}$
        &$\bm{70.65\pm{0.33}}$
        &$\bm{85.76\pm{0.59}}$
        \\
        \Xhline{2\arrayrulewidth}

        DeepEMD*$^\clubsuit$~\cite{zhang2020deepemd}      
        &ResNet-12
        &$67.37\pm{0.45}$                       
        &$83.17\pm{0.75}$  
        &$73.19\pm{0.32}$                               
        &$86.79\pm{0.61}$  
        \\
       \rowcolor{Gray}
        \textbf{INSTA-DeepEMD*$^\clubsuit$}   
        &ResNet-12
        &$\bm{68.46\pm{0.48}}$                      
        &$\bm{84.21\pm{0.82}}$
        &$\bm{73.87\pm{0.31}}$                    
        &$\bm{88.02\pm{0.61}}$
        \\

    \Xhline{2\arrayrulewidth}
    \end{tabular}
    }
    \end{center}
    \label{tab:mini-tiered}
\end{table}

\noindent{\bf{\emph{mini}-ImageNet.}}
As shown in Table~\ref{tab:mini-tiered}, our method improves the performance of all the baseline models by a noticeable margin. It is worthwhile to mention that our INSTA can boost the performance of the baseline model ProtoNet by $4.72\%$ and $3.67\%$, and achieves $67.01\%$ and $83.13\%$ for 5-way 1-shot and 5-way 5-shot settings, which outperforms many recent published models. Furthermore, we can show an improvement over a strong baseline model, \ie DeepEMD, and achieve state-of-the-art performance on this dataset. We provide more comparison between our approach and other state-of-the-art methods in our supplementary material for \emph{mini}-ImageNet and \emph{tiered}-ImageNet.

\noindent{\bf{\emph{tiered}-ImageNet.}}
As shown in Table~\ref{tab:mini-tiered}, our model consistently brings the performance gain over baseline models. Among baseline models, our model shows a greater performance improvement over MAML than ProtoNet and DeepEMD. Notably, a significant improvement on 5-way 5-shot can be seen over DeepEMD. We attribute this improvement to our model's context module leveraging the categorical nature of this dataset.
The results also show that the DeepEMD exhibits a performance gain with our design and outperforms many other recent models, which achieves the state-of-the-art result on this dataset.

\begin{table}[!h]

     \caption{Few-shot classification accuracy and 95\% confidence interval on CUB and FC100 with ResNet backbones 
}
    \begin{center}
    \scalebox{0.9}{
    \begin{tabular}{l c| c c| c c}
        \Xhline{2\arrayrulewidth}
         \multirow{2}{*}{Model}                      
         &\multirow{2}{*}{Backbone}
         &\multicolumn{2}{c|}{CUB}
         &\multicolumn{2}{c}{FC100}
         \\
         
         &
         &{5-way 1-shot}              &{5-way 5-shot}
         &{5-way 1-shot}              &{5-way 5-shot}
        \\
    \Xhline{2\arrayrulewidth}
        
        RelationNet~\cite{sung2018learning}
        &ResNet-18
        &$67.59\pm{1.02}$                    
        &$82.75\pm{0.58}$
        &-
        &-
        \\
        Chen~\etal~\cite{chen2019closer}
        &ResNet-18
        &$67.02$                    
        &$83.58$
        &-
        &-
        \\
        SimpleShot~\cite{wang2019simpleshot}
        &ResNet-18
        &$70.28$                    
        &$86.37$
        &-
        &-
        \\
        Neg-Margin~\cite{liu2020negative}
        &ResNet-18  
        &$72.66\pm{0.85}$
        &$89.40\pm{0.43}$
        &-
        &-
        
        \\
        P-transfer~\cite{shen2021partial}
        &ResNet-12
        &$73.88\pm{0.87}$                    
        &$87.81\pm{0.48}$
        &-
        &-
        \\
         TADAM~\cite{oreshkin2018tadam}      
        &ResNet-12
        &-
        &-
        &$40.10\pm{0.40}$                   
        &$56.10\pm{0.40}$    
        \\
        ConstellationNet~\cite{xu2020attentional}              
        &ResNet-12
        &-
        &-
        &$43.80\pm{0.20}$                   
        &$59.70\pm{0.20}$    
        \\
       
        \Xhline{2\arrayrulewidth}
        
        MAML*~\cite{finn2017model}    
        &ResNet-10
        &$70.46\pm{0.97}$          
        &$80.15\pm{0.73}$ 
        &$34.50\pm{0.69}$
        &$47.31\pm{0.68}$
        
        \\
        \rowcolor{Gray}
        \textbf{INSTA-MAML*}
        &ResNet-10
        &$\bm{73.08\pm{0.97}}$
        &$\bm{84.26\pm{0.66}}$
        &$\bm{38.02\pm{0.70}}$
        &$\bm{51.20\pm{0.68}}$
        \\
        \Xhline{2\arrayrulewidth}
        
        ProtoNet*~\cite{snell2017prototypical}    
        &ResNet-12
        &$64.76\pm{0.33}$          
        &$77.99\pm{0.68}$ 
        &$41.54\pm{0.76}$
        &$57.08\pm{0.76}$
        
        \\
        \rowcolor{Gray}
        \textbf{INSTA-ProtoNet*}
        &ResNet-12
        &$\bm{69.05\pm{0.32}}$
        &$\bm{81.71\pm{0.63}}$
        &$\bm{44.12\pm{0.26}}$
        &$\bm{62.04\pm{0.75}}$
        \\
        
        \Xhline{2\arrayrulewidth}

        DeepEMD*$^\clubsuit$~\cite{zhang2020deepemd}           
        &ResNet-12
        &$74.55\pm{0.30}$                       
        &$87.55\pm{0.54}$   
        &$45.12\pm{0.26}$                   
        &$61.46\pm{0.70}$    
        \\
        \rowcolor{Gray}
        \textbf{INSTA-DeepEMD*$^\clubsuit$}
        &ResNet-12
        &$\bm{75.26\pm{0.31}}$              
        &$\bm{88.12\pm{0.54}}$
        &$\bm{45.42\pm{0.26}}$               
        &$\bm{62.37\pm{0.68}}$
        \\
        \Xhline{2\arrayrulewidth}
    \end{tabular}
    }
    \end{center}
    \label{tab:CUB-FC100}
\end{table}

\noindent{\bf{CUB.}} The result on the CUB dataset is shown in Table~\ref{tab:CUB-FC100}, where the performance of all the three baseline models is improved by integrating the proposed INSTA. For the ProtoNet baseline with ResNet-12 backbone, as an example, $4.29\%$ and $3.72\%$ improvements can be observed for 5-way 1-shot and 5-way 5-shot, respectively. Moreover, among the improved models, our INSTA-DeepEMD achieves the state-of-the-art results, which are $75.26\%$ and $88.12\%$ for 5-way 1-shot and 5-way 5-shot settings, respectively. Please refer to supplementary material for additional results of the INSTA-ProtoNet with ResNet-18 backbone.

\noindent{\bf{FC100.}} 
Consistent with the observation in the other benchmarks, the results in Table~\ref{tab:CUB-FC100} again show that a performance improvement can be achieved on FC100. Furthermore, both INSTA-ProtoNet and INSTA-DeepEMD achieve comparable performance with the recent state-of-the-art models on FC100, which vividly shows the effectiveness of our approach.

\subsection{Few-Shot Detection}\label{ssec:fsd}
\noindent{\bf{Problem Definition.}} 
Given a query and several support images (each support image only contains one object), a few-shot detection task is to output labels and corresponding bounding boxes for all objects in the query image that belong to support categories. Specifically, few-shot detection follows the N-way K-shot setting, where the support set contains N categories, with each category having K samples. In the following, we will discuss the experiment setup of this task. The implementation details on incorporating our INSTA with the baseline model proposed by Fan~\etal~\cite{fan2020few} are included in the supplementary material. 

\noindent{\bf{Experiment Setup.}} 
In this experiment, we follow the training and testing settings in~\cite{fan2020few}, where a 2-way 9-shot contrastive training strategy is adopted, and a 20-way 10-shot setting is used for final testing. 


\noindent{\bf{Results.}}
\begin{table}[!h]
     \caption{The few-shot detection results of 6 different average precision (AP) on the test set, including MS COCO and PASCAL VOC datasets with 20-way 10-shot setting}
    \begin{center}
    \scalebox{1}{
    \begin{tabular}{l|c|c|c|c|c|c}
        \Xhline{2\arrayrulewidth}
         {Model}                      &{AP}
         &{AP$_{50}$}              &{AP$_{75}$}     
         &{AP$_{s}$}
         &{AP$_{m}$}
         &{AP$_{l}$}
        \\
       \Xhline{2\arrayrulewidth}
        FR~\cite{kang2019few}    
        &$5.6$
        &$12.3$
        &$4.6$
        &-
        &-
        &-
        \\
        Meta~\cite{yan2019meta}
        &$8.7$
        &$19.1$
        &$6.6$
        &-
        &-
        &-
        \\
        Fan~\etal~\cite{fan2020few}
        &$11.1$
        &$20.4$
        &$10.6$
        &$2.8$
        &$12.3$
        &$20.7$
    
        \\
        \rowcolor{Gray}
        \textbf{Fan~\etal+INSTA}
        &$\bm{12.5}$
        &$\bm{23.6}$
        &$\bm{12.1}$
        &$\bm{3.3}$
        &$\bm{13.2}$
        &$\bm{21.4}$
        \\
      
    \Xhline{2\arrayrulewidth}
    \end{tabular}
    }
    \end{center}
    \label{tab:FSD}
\end{table}

We compare our model against the baseline model and other relevant state-of-the-art few-shot detection models. As shown in Table~\ref{tab:FSD}, by incorporating the proposed INSTA on the baseline model, improved performance can be seen across all of the metrics. Specifically, as compared to the baseline model, the AP$_{50}$ is improved by $3.2\%$ and achieves $23.6\%$, which empirically shows that our dynamic kernels indeed extract more informative and representative features. Moreover, the proposed INSTA improves the performance of detection for objects of all the scales (refer to AP$_{s}$, AP$_{m}$, and AP$_{l}$), which again illustrates the importance of our context module when adapting to different object scales in a given task. 

\subsection{Ablation Study}\label{subsec:ablation}
In the following section, we conduct ablation studies to discuss and verify the effect of each component of our framework, including the instance-kernel, task-specific kernel, MSA \vs GAP, and we also evaluate some variants of the proposed INSTA to justify the selection of our current framework. In this study, we use the ProtoNet as the baseline, and the ResNet-12 is adopted as its backbone. Additionally, this study is conducted on the \emph{mini}-ImageNet under the 5-way 5-shot setting. The results are summarized in Table~\ref{tab:ablation}.   
\begin{table}[!h]
    \caption{The ablation study of each component in our framework}
    \begin{center}
    \scalebox{1}{
    \begin{tabular}{l|c|c|c|c}
       \Xhline{2\arrayrulewidth}
       {ID}
       &{Model}
       &{Apply to $\Mat{S}_{ij}$}
       &{Apply to $\Mat{Q}$}
       &{5-way 5-shot}     
       \\
       \Xhline{2\arrayrulewidth}
       (i)
       &ProtoNet
       &-
       &-
       &$79.46\pm{0.48}$

       \\
       (ii)
       &ProtoNet + $\Mat{G}^{ta}$
       &$\Mat{G}^{ta}$ 
       &-
       &$81.56\pm{0.57}$
        \\
       
       (iii)
       &ProtoNet + $\Mat{G}^{ta}$
       &$\Mat{G}^{ta}$
       &$\Mat{G}^{ta}$
       & $82.51\pm{0.58}$   
       \\
       (iv)
       &ProtoNet + $\Mat{G}^{in}$
       &$\Mat{G}^{in}$
       &-
       &$80.81\pm{0.60}$
       
       \\
       
       (v)
       &ProtoNet + $\Mat{G}^{insta}$
       &$\Mat{G}^{ta}, \Mat{G}^{in}$
       &-
       & $81.74\pm{0.56}$
       
       \\    
      
      (vi)
      &INSTA-ProtoNet + GAP
      &$\Mat{G}^{ta}, \Mat{G}^{in}$
      &$\Mat{G}^{ta}$
      & $82.07\pm{0.56}$    
      \\

      (vii)
      &INSTA-ProtoNet + $\Mat{G}^{in}_{\Mat{Q}}$
      &$\Mat{G}^{ta}, \Mat{G}^{in}$
      &$\Mat{G}^{in}_{\Mat{Q}}, \Mat{G}^{ta}$
      & $82.24\pm{0.56}$    
      \\    
       
      (viii)
      &INSTA-ProtoNet w/o sharing $f_{\omega}$ 
      &$\Mat{G}^{ta}, \Mat{G}^{in}$
      &$\Mat{G}^{ta}$
      &$81.36\pm{0.59}$    
      \\    
       \rowcolor{Gray}
       
      (ix)
      &INSTA-ProtoNet 
      &$\Mat{G}^{ta}, \Mat{G}^{in}$
      &$\Mat{G}^{ta}$
      &$\bm{83.13\pm{0.56}}$    
      \\    
      
    \Xhline{2\arrayrulewidth}
    \end{tabular}
    }
    \end{center}
    \label{tab:ablation}
\end{table}

\noindent{\bf{Effectiveness of Task-Specific Kernel.}}  We study the effect of the task-specific kernel in this experiment (\ie, (ii) in Table~\ref{tab:ablation}), where we only apply the task-specific kernel to support samples and leave the query unchanged. By comparing (i) and (ii), we can observe that our task-specific kernel improves the performance of ProtoNet by $2.10\%$. Moreover, we apply the task-specific kernel on the query in setting (iii), and we can observe a further improvement over the setting (ii), which verifies the effectiveness of our task-specific kernel.

\noindent{\bf{Effectiveness of Instance Kernel.}} In setting (iv), we enable the instance kernel based on ProtoNet (\ie, (i)) and the performance of ProtoNet is improved by $1.35\%$. Moreover, comparing setting (ii), where the only task-specific kernel is enabled, to setting (v), where both task-specific and instance kernels are enabled, a performance improvement can also be observed. Both cases illustrate the effectiveness of our instance kernel.

\noindent{\bf{Effectiveness of query adaptation.}} The purpose of this experiment is to study the effect of the adaptation on the query feature. In setting (iii), we enable the task-specific kernel on both support and query features but disable the instance kernel. Compared to setting (ii), where the task-specific kernel is enabled only on the support features, setting (iii) yields a better performance. Furthermore, the comparison between setting (v) and (ix) highlights the importance of adapting the query feature map using the task information for FSL tasks.

\noindent{\bf{Effectiveness of MSA.}} In this study, we show the performance gap between using GAP and MSA (\S~\ref{para:channel_branch}). In setting (vi) we replace the MSA in our final design (ix) with GAP. As the results in Table~\ref{tab:ablation} showed, the MSA indeed helps with learning a more informative and representative dynamic kernel than GAP.

\noindent{\bf{Instance Kernel for Query.}} In our framework, we only use task-specific kernels on the query feature map since the instance kernels obtained from support samples might not be instance level representative for the query sample but provide useful task information. Therefore, we perform extra experiments to verify whether the instance kernel obtained from the query itself can extract better features. As the result in (vii) indicates, the instance kernel extracted from the query feature map does not improve the performance, as compared to our final design. 

\noindent{\bf{Shared Dynamic Kernel Generator.}} We hypothesize that sharing the Dynamic Kernel Network for the task-specific kernel and instance kernel encourages learning a more representative instance kernel. To verify this, we further conduct an experiment where two independent dynamic kernel generators are used to produce the task-specific kernel and instance kernel. As the comparison between (ix) and (viii) shows, inferring the instance and task kernels using a shared weight dynamic kernel generator is an essential design choice.

%% file: Conclusion/conclusion.tex
\section{Conclusion}
In this paper, we propose to learn a dynamic embedding function realized by a novel dynamic kernel, which extracts features at both instance-level and task-level while encoding important frequency patterns. Our method improves the performance of several FSL models. This is demonstrated on 4 public few-shot classification datasets, including \emph{mini}-ImageNet, \emph{tiered}-ImageNet, CUB, and FC100 and a few-shot detection dataset, namely MS COO-PASCAL-VOC. 


%% file: Appendix/appendix.tex
\section{Implementation Details}
\subsection{Datasets}
\noindent{\bf{\emph{mini}-ImageNet.}} The \emph{mini}-ImageNet is sampled from ImageNet~\cite{deng2009imagenet}. This dataset has 100 classes, with each having 600 samples. We follow the standard protocol~\cite{Ravi2017OptimizationAA} to split the dataset into 64 training, 16 validation, and 20 testing classes.

\noindent{\bf{\emph{tiered}-ImageNet.}} Similar to \emph{mini}-ImageNet, \emph{tiered}-ImageNet is also a subset of the ImageNet. This dataset consists of 608 classes from 34 categories and is split into 351 classes from 20 categories for training, 97 classes from 6 categories for validation, and 160 classes from 8 categories for testing. 

\noindent{\bf{CUB.}} The CUB is a fine-grained dataset, which consists of 11,788 images from 200 different breeds of birds. We follow the standard settings~\cite{liu2020negative}, in which the dataset is split into 100/50/50 breeds for training, validation, and testing, respectively.

\noindent{\bf{FC100.}} FC100 dataset is a variant of the standard CIFAR100 dataset~\cite{krizhevsky2009learning}, which contains images from 100 classes, with each class containing 600 samples. We follow the standard setting~\cite{oreshkin2018tadam}, where the dataset is split into 60/20/20 classes for training, validation and testing, respectively.

\noindent{\bf{MS COCO and PASCAL VOC Datasets.}} In the few-shot detection task, we follow the protocol used in~\cite{fan2020few} to construct the dataset, where images from 60 categories of the MS COCO dataset are used for training and images from the rest of 20 common categories between MS COCO and PASCAL VOC datasets are used for testing. 

\subsection{Few-Shot Classification Hyperparameters}

\noindent{\textbf{Network and Optimizer.}} 
We use the ResNet-10 backbone~\cite{chen2019closer} for the MAML and the ResNet-12 backbone~\cite{ye2020few, zhang2020deepemd} for the other two baselines across all four benchmarks. Noted additional ResNet-18 back1bone~\cite{ziko2020laplacian} is employed for the ProtoNet experiments on CUB. We fix the size of input images to $84\times 84$ for ProtoNet and DeepEMD baselines and $224\times 224$ for MAML baseline (We strictly follow the same pre-processing protocol\footnote{\href{https://github.com/icoz69/DeepEMD}{}\url{https://github.com/icoz69/DeepEMD}.} in the original DeepEMD implementation to implement the metric-based baselines and our model, \ie, DeepEMD, ProtoNet, INSTA-DeepEMD, and INSTA-ProtoNet. For MAML, we strictly follow the pre-processing protocol implemented in~\cite{chen2019closer}\footnote{\href{https://github.com/wyharveychen/CloserLookFewShot}{}\url{https://github.com/wyharveychen/CloserLookFewShot}.}). We use SGD optimizer for ProtoNet and DeepEMD experiments~\cite{ye2020few,zhang2020deepemd} and AdamW optimizer for MAML experiments~\cite{chen2019closer} across all the datasets. For ProtoNet and DeepEMD baselines, we use L2 regularizer with 0.0005 weight decay factor. In the MAML baseline, the weight dacay factor is 0.01. For ResNet-18 and ResNet-10 backbones, we disable the average pooling and remove the last fully connected (FC) layer to produce the feature maps with size of $512\times 11\times 11$ and $512\times 7\times 7$, respectively. In the ResNet-12 backbone, the network produces the feature map with a size of $640\times 5\times 5$.
\\\noindent{\textbf{Training.}} We follow the good practice in the state-of-the-art models~\cite{simon2020adaptive, zhang2020deepemd, ye2020few}, where the network training is split into two stages, \ie pre-training and meta-training stages. During pre-training stage, the backbone network with a FC layer is trained on all the training classes of the dataset with the standard classification task. We select the network with the highest validation accuracy as the pre-trained backbone network for the meta-training stage. During meta-training stage, we follow the standard episodic training protocol~\cite{vinyals2016matching} to train the entire model. We set a small learning rate ($0.0002$) for the backbone and a larger learning rate ($0.0002\times 25$) for the other modules during meta-training stage.  Additionally, we use cosine annealing learning rate scheduler over 200 epochs.

\begin{figure}[!h]
\begin{center}
   \scalebox{1.1}{
   \includegraphics[width=0.9\linewidth]{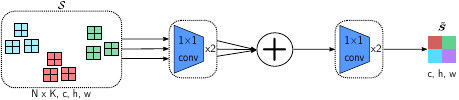}
}
\end{center}
   \caption{The conceptual diagram of the context learning module
   } 
\label{fig:contex}
\end{figure}

\subsection{Few-Shot Detection}
For the few-shot detection task, we consider the method proposed by Fan~\etal~\cite{fan2020few} as our baseline model, which inherits from the faster-R-CNN~\cite{ren2015faster} framework. Similar to the few-shot classification task, we implement our method to produce dynamic kernels and perform convolution on the feature maps extracted by the backbone network (\ie ResNet-50), which is followed by a region proposal network (RPN), a region of interest pooling (ROI pooling) operation, and the classification and bounding box regression heads, outputting the categories and bounding boxes for the objects to the query image. Notably, For a fair comparison, we do not use any additional data augmentation. Please refer to~\cite{fan2020few} for more details of the framework and training strategy.

\section{Additional Experiment Results}
\subsection{\emph{mini}-ImageNet, \emph{tiered}-ImageNet, and CUB} 
In this part, we provide extra comparison between our model and other state-of-the-art models on \emph{mini}-ImageNet, \emph{tiered}-ImageNet, and CUB. We follow the same evaluation protocol in~\cite{zhang2020deepemd}\footnote{\href{https://github.com/icoz69/DeepEMD/blob/master/eval.py}{}\url{https://github.com/icoz69/DeepEMD/blob/master/eval.py}.} to evaluate the metric-based baseline models and our models (\ie, DeepEMD, ProtoNet, INSTA-DeepEMD, and INSTA-ProtoNet), where 5,000 and 600 episodes are randomly sampled for 1-shot and 5-shot settings. For MAML baseline and INSTA-MAML, we strictly follow the same evaluation protocol in~\cite{chen2019closer}\footnote{\href{https://github.com/wyharveychen/CloserLookFewShot/blob/master/test.py}{}\url{https://github.com/wyharveychen/CloserLookFewShot/blob/master/test.py}.},
where 600 episodes are randomly sampled for both 1-shot and 5-shot settings.
\begin{table}[!h]
\caption{Few-shot classification accuracy and 95\% confidence interval on \emph{mini}-ImageNet and \emph{tiered}-ImageNet with ResNet backbones 
    }
    \begin{center}
    \scalebox{0.9}{
    \begin{tabular}{l c| c c|c c}
        \Xhline{2\arrayrulewidth}
         \multirow{2}{*}{Model}                      &\multirow{2}{*}{Backbone}
         &\multicolumn{2}{c|}{\emph{mini}-ImageNet}
         &\multicolumn{2}{c}{\emph{tiered}-ImageNet}
         \\
         &
         &{5-way 1-shot}              &{5-way 5-shot}
         &{5-way 1-shot}              &{5-way 5-shot}     
        \\
        \Xhline{2\arrayrulewidth}

        
        MetaOptNet-SVM~\cite{lee2019meta}            
        &ResNet-12
        &$64.09\pm{0.62}$                      
        &$80.00\pm{0.45}$
        &-
        &-
        \\
        Neg-Margin~\cite{liu2020negative}           
        &ResNet-12
        &$63.85\pm{0.81}$                      
        &$81.57\pm{0.56}$
        &-
        &-
        
        \\
        \Xhline{2\arrayrulewidth}
        TPN~\cite{liu2018learning}                         
        &ResNet-12
        &$59.46$                      
        &$75.65$
        &-
        &-
        \\
        DSN-MR~\cite{simon2020adaptive}             
        &ResNet-12
        &$64.60\pm{0.72}$                      
        &$79.51\pm{0.50}$
        &$67.39\pm{0.82}$                   
        &$82.82\pm{0.56}$
        \\
        E$^3$BM~\cite{liu2020ensemble}              
        &ResNet-12
        &$63.80\pm{0.40}$                      
        &$80.10\pm{0.30}$
        &$71.20\pm{0.40}$                      
        &$85.30\pm{0.30}$
        
        
        \\
        ConstellationNet~\cite{xu2020attentional}        
        &ResNet-12
        &$64.89\pm{0.23}$       
        &$79.95\pm{0.37}$
        &-
        &-
        \\
    

        MELR~\cite{fei2020melr}                    
        &ResNet-12
        &$67.40\pm{0.43}$       
        &$83.40\pm{0.28}$
        &$72.14\pm{0.51}$                      
        &$87.01\pm{0.35}$
        \\
        CNL~\cite{zhao2021looking}                     
        &ResNet-12
        &$67.96\pm{0.98}$       
        &$83.36\pm{0.51}$
        &$73.42\pm{0.95}$                      
        &$87.72\pm{0.75}$
        \\
        
        \Xhline{2\arrayrulewidth}
        
        MAML*~\cite{finn2017model} 
        &ResNet-10
        &$54.73\pm{0.87}$
        &$66.72\pm{0.81}$  
        &$59.85\pm{0.97}$                   
        &$73.20\pm{0.81}$
        \\
        
        \rowcolor{Gray}
        \textbf{INSTA-MAML*}  
        &ResNet-10
        &$\bm{56.41\pm{0.87}}$
        &$\bm{71.56\pm{0.75}}$
         &$\bm{63.34\pm{0.92}}$                    
        &$\bm{78.01\pm{0.71}}$
        
        \\
        \Xhline{2\arrayrulewidth}
        ProtoNet*~\cite{snell2017prototypical}                    
        &ResNet-12
        &$62.29\pm{0.33}$            
        &$79.46\pm{0.48}$ 
        &$68.25\pm{0.23}$                   
        &$84.01\pm{0.56}$
      
       
        \\ 
        \rowcolor{Gray}
        \textbf{INSTA-ProtoNet*}              
        &ResNet-12
        &$\bm{67.01\pm{0.30}}$                      &$\bm{83.13\pm{0.56}}$
        &$\bm{70.65\pm{0.33}}$
        &$\bm{85.76\pm{0.59}}$
        \\

        DeepEMD*$^\clubsuit$~\cite{zhang2020deepemd}      
        &ResNet-12
        &$67.37\pm{0.45}$                       
        &$83.17\pm{0.75}$
        &$73.19\pm{0.32}$                               
        &$86.79\pm{0.61}$
        \\
       \rowcolor{Gray}
        \textbf{INSTA-DeepEMD*$^\clubsuit$}   
        &ResNet-12
        &$\bm{68.46\pm{0.48}}$                      
        &$\bm{84.21\pm{0.82}}$   
        &$\bm{73.87\pm{0.31}}$                    
       &$\bm{88.02\pm{0.61}}$
        \\

    \Xhline{2\arrayrulewidth}
    \end{tabular}
    }
    \end{center}
    \label{tab:mini}
\end{table}

\begin{table}[!h]
\caption{Few-shot classification accuracy and 95\% confidence interval on CUB with ResNet backbones
    }
    \begin{center}
    \scalebox{0.8}{
    \begin{tabular}{l c| c c}
        \Xhline{2\arrayrulewidth}
         {Model}                      &{Backbone}
         &{5-way 1-shot}              &{5-way 5-shot}     
        \\
   
        \Xhline{2\arrayrulewidth}
        
        MAML*~\cite{finn2017model}    
        &ResNet-10
        &$70.46\pm{0.97}$          
        &$80.15\pm{0.73}$ 
        
        \\
        \rowcolor{Gray}
        \textbf{INSTA-MAML*}
        &ResNet-10
        &$\bm{73.08\pm{0.97}}$
        &$\bm{84.26\pm{0.66}}$
        \\
       
        \Xhline{2\arrayrulewidth}

        DeepEMD*$^\clubsuit$~\cite{zhang2020deepemd}           
        &ResNet-12
        &$74.55\pm{0.30}$                       
        &$87.55\pm{0.54}$    
        \\
        \rowcolor{Gray}
        \textbf{INSTA-DeepEMD*$^\clubsuit$}
        &ResNet-12
        &$\bm{75.26\pm{0.31}}$              
        &$\bm{88.12\pm{0.54}}$
        \\
        
        \Xhline{2\arrayrulewidth}

        ProtoNet*    
        &ResNet-18
        &$75.06\pm{0.30}$          
        &$87.39\pm{0.48}$ 
        \\
        \rowcolor{Gray}
        \textbf{INSTA-ProtoNet*}
        &ResNet-18
        &$\bm{77.18\pm{0.29}}$
        &$\bm{89.54\pm{0.44}}$
        \\
        \Xhline{2\arrayrulewidth}
    \end{tabular}
    }
    \end{center}
    
    \label{tab:CUB}
\end{table}

\subsection{Meta-Dataset}
\begin{table}[!h]
    \caption{Few-shot classification results on Meta-dataset with ResNet-18 backbone}
    \begin{center}
    \begin{tabular}{l|c | c}
       \Xhline{2\arrayrulewidth}
        {Dataset}
        &{Simple-CNAPS}
        &{INSTA-Simple-CNAPS}
        \\
        \Xhline{2\arrayrulewidth}
        ILSVRC
        &$55.5\pm{1.1}$
        &$\bm{58.5\pm{1.1}}$
        \\
        Omniglot
        &$91.0\pm{0.6}$
        &$\bm{91.9\pm{0.6}}$
        \\
        Aircraft
        &$81.2\pm{0.7}$
        &$\bm{82.4\pm{0.8}}$
        \\
        Birds
        &$74.3\pm{0.9}$
        &$\bm{75.7\pm{0.8}}$
        \\
        Textures
        &$66.9\pm{0.8}$
        &$\bm{67.8\pm{0.8}}$
        \\
        Quick Draw
        &$76.7\pm{0.8}$
        &$\bm{76.8\pm{0.8}}$
        \\
        Fungi
        &$47.5\pm{1.0}$
        &$\bm{49.2\pm{1.1}}$
        \\
        VGG Flowers
        &$\bm{90.5\pm{0.6}}$
        &$90.4\pm{0.6}$
        \\
        Traffic Signs
        &$72.0\pm{0.7}$
        &$\bm{74.1\pm{0.7}}$
        \\
        MSCOCO
        &$47.3\pm{1.1}$
        &$\bm{53.9\pm{1.1}}$
        \\
       \Xhline{2\arrayrulewidth}
    \end{tabular}
    \end{center}
 
    \label{tab:meta}
\end{table}

To verify the effectiveness of our method on the cross-domain few-shot classification problem, we incorporate INSTA into the baseline model simple-CNAPS~\cite{Bateni2020_SimpleCNAPS}. In this experiment, we fix the trained baseline model and only fine-tune the modules to generate the INSTA dynamic kernels (\ie, dynamic kernel generator and context learning module). We follow the same implementation of simple-CNAPS\footnote{\href{https://github.com/peymanbateni/simple-cnaps/tree/master/simple-cnaps-src}{}\url{https://github.com/peymanbateni/simple-cnaps/tree/master/simple-cnaps-src}.} to conduct this experiment (\eg, $8\times 10^{-3}$ as learning rate, Adam as the optimizer, \etc). As the results in Table~\ref{tab:meta} suggested, INSTA improves the baseline over almost all the datasets, which again shows the effectiveness of our proposed INSTA dynamic kernels.

\subsection{Ablation Study}
In this part, we provide extra ablation studies on the effect of the residual connection and the spatial size of our dynamic kernel.

\noindent{\textbf{Residual Connection}.}
In this experiment, we study the effect of the residual connection in our framework. In setting (i) of the Table~\ref{tab:extra_a}, we disable the residual connection between the adapted and original features. The result suggests that the residual connection is an essential design choice for our framework.

\noindent{\textbf{Kernel Size}.}
We provide an extra study on the effect of the spatial size of our dynamic kernel. Given that the feature map extracted by ResNet-12 has spatial size $5\times 5$, the dynamic kernel size is constrained smaller or equal to $5\times 5$. Therefore, in this study, we compare the results when the dynamic kernel size $k=5\times 5$
to our final design ($k=3\times 3$).

\begin{table}[!h]
\caption{The extra ablation study for the effect of the residual connection and spatial size of our dynamic kernel}
    \begin{center}
    \scalebox{1}{
    \begin{tabular}{c|c|c}
       \Xhline{2\arrayrulewidth}
      
       ID
       &Model
       &{5-way 5-shot}     
       \\
       \Xhline{2\arrayrulewidth}
       (i)
       &INSTA-ProtoNet w/o residual
       &$80.03$ 
       \\
       (ii)
       &INSTA-ProtoNet w/ $5\times 5~\Mat{G}^{dy}$ 
       &$82.43$
      \\    
      \rowcolor{Gray}
      (iii)
      &INSTA-ProtoNet 
      &$83.13$
      \\
    \Xhline{2\arrayrulewidth}
    \end{tabular}
    }
    \end{center}

    \label{tab:extra_a}
\end{table}

\section{Multi-Spectral Attention}
In this section, we provide more details for using the 2D-DCT to obtain the frequency-encoded vector. We first introduce the basis function of the 2D-DCT. The basis $B_{u,v}^{a,b}$ of the 2D-DCT is given by:
\begin{equation}\label{eq:basis}
B_{u,v}^{a,b} = \mathrm{cos}(\frac{\pi u}{h}(a+\frac{1}{2}))\mathrm{cos}(\frac{\pi v}{w}(b+\frac{1}{2})),
\end{equation}
where $u, v$ are the frequency components of a basis. Then the frequency-encoded vector of a 3D-tensor $\Mat{S}^i\in\mathbb{R}^{\frac{c}{n}\times h\times w}$ can be obtained by:
\begin{gather*}\label{eq:DCT}
\Vec{\tau}^i = \sum_{a=0}^{h-1}\sum_{b=0}^{w-1}\Mat{S}_{:,a,b}B_{u_i,v_i}^{a,b} 
\\ 
\st~i\in\{0,1,\ldots, n-1\}, 
\end{gather*}
where $\Vec{\tau}^i\in\mathbb{R}^{\frac{c}{n}}$ is the frequency-encoded vector, $h$ and $w$ are the height and width of the input signal. We pick the lowest 16 frequency components for our basis function according to~\cite{qin2020fcanet}. Finally, we concatenate all the frequency-encoded vectors as: 
\begin{equation}\label{eq:cat_supp}
 \Mat{\tau} = \mathrm{concat}(\Mat{\tau}^0,\Mat{\tau}^1,\ldots,\Mat{\tau}^{n-1}),
\end{equation}
where $\Vec{\tau}\in\mathbb{R}^{c}$.

%% file: eccv2022submission.bbl
\begin{thebibliography}{10}
\providecommand{\url}[1]{\texttt{#1}}
\providecommand{\urlprefix}{URL }
\providecommand{\doi}[1]{https://doi.org/#1}

\bibitem{andrychowicz2016learning}
Andrychowicz, M., Denil, M., Gomez, S., Hoffman, M.W., Pfau, D., Schaul, T.,
  Shillingford, B., De~Freitas, N.: Learning to learn by gradient descent by
  gradient descent. In: Advances in neural information processing systems. pp.
  3981--3989 (2016)

\bibitem{antoniou2019train}
Antoniou, A., Edwards, H., Storkey, A.: How to train your maml. In:
  International Conference on Learning Representations (2019)

\bibitem{Bateni2020_SimpleCNAPS}
Bateni, P., Goyal, R., Masrani, V., Wood, F., Sigal, L.: Improved few-shot
  visual classification. In: Proceedings of the IEEE/CVF Conference on Computer
  Vision and Pattern Recognition (CVPR) (June 2020)

\bibitem{bertinetto2018meta}
Bertinetto, L., Henriques, J.F., Torr, P., Vedaldi, A.: Meta-learning with
  differentiable closed-form solvers. In: International Conference on Learning
  Representations (2018)

\bibitem{bertinetto2016learning}
Bertinetto, L., Henriques, J.F., Valmadre, J., Torr, P.H., Vedaldi, A.:
  Learning feed-forward one-shot learners. In: Proceedings of the 30th
  International Conference on Neural Information Processing Systems. pp.
  523--531 (2016)

\bibitem{bolukbasi2017adaptive}
Bolukbasi, T., Wang, J., Dekel, O., Saligrama, V.: Adaptive neural networks for
  efficient inference. In: International Conference on Machine Learning. pp.
  527--536. PMLR (2017)

\bibitem{chen2019closer}
Chen, W.Y., Liu, Y.C., Kira, Z., Wang, Y.C.F., Huang, J.B.: A closer look at
  few-shot classification. arXiv preprint arXiv:1904.04232  (2019)

\bibitem{chen2020dynamic}
Chen, Y., Dai, X., Liu, M., Chen, D., Yuan, L., Liu, Z.: Dynamic convolution:
  Attention over convolution kernels. In: Proceedings of the IEEE/CVF
  Conference on Computer Vision and Pattern Recognition. pp. 11030--11039
  (2020)

\bibitem{choi2018structured}
Choi, J., Krishnamurthy, J., Kembhavi, A., Farhadi, A.: Structured set matching
  networks for one-shot part labeling. In: Proceedings of the IEEE Conference
  on Computer Vision and Pattern Recognition. pp. 3627--3636 (2018)

\bibitem{deng2009imagenet}
Deng, J., Dong, W., Socher, R., Li, L.J., Li, K., Fei-Fei, L.: Imagenet: A
  large-scale hierarchical image database. In: 2009 IEEE conference on computer
  vision and pattern recognition. pp. 248--255. Ieee (2009)

\bibitem{fan2020few}
Fan, Q., Zhuo, W., Tang, C.K., Tai, Y.W.: Few-shot object detection with
  attention-rpn and multi-relation detector. In: Proceedings of the IEEE/CVF
  Conference on Computer Vision and Pattern Recognition. pp. 4013--4022 (2020)

\bibitem{fang2021kernel}
Fang, P., Harandi, M., Petersson, L.: Kernel methods in hyperbolic spaces. In:
  Proceedings of the IEEE/CVF International Conference on Computer Vision. pp.
  10665--10674 (2021)

\bibitem{AIA_Pengfei_PAMI}
Fang, P., Zhou, J., Roy, S.K., Ji, P., Petersson, L., Harandi, M.: Attention in
  attention networks for person retrieval. IEEE Transactions on Pattern
  Analysis and Machine Intelligence pp.~1--1 (2021)

\bibitem{fei2020melr}
Fei, N., Lu, Z., Xiang, T., Huang, S.: Melr: Meta-learning via modeling
  episode-level relationships for few-shot learning. In: International
  Conference on Learning Representations (2020)

\bibitem{finn2017model}
Finn, C., Abbeel, P., Levine, S.: Model-agnostic meta-learning for fast
  adaptation of deep networks. arXiv preprint arXiv:1703.03400  (2017)

\bibitem{flennerhag2019meta}
Flennerhag, S., Rusu, A.A., Pascanu, R., Visin, F., Yin, H., Hadsell, R.:
  Meta-learning with warped gradient descent. arXiv preprint arXiv:1909.00025
  (2019)

\bibitem{franceschi2018bilevel}
Franceschi, L., Frasconi, P., Salzo, S., Grazzi, R., Pontil, M.: Bilevel
  programming for hyperparameter optimization and meta-learning. arXiv preprint
  arXiv:1806.04910  (2018)

\bibitem{ha2016hypernetworks}
Ha, D., Dai, A., Le, Q.V.: Hypernetworks. arXiv preprint arXiv:1609.09106
  (2016)

\bibitem{hou2019cross}
Hou, R., Chang, H., Ma, B., Shan, S., Chen, X.: Cross attention network for
  few-shot classification. arXiv preprint arXiv:1910.07677  (2019)

\bibitem{hu2018squeeze}
Hu, J., Shen, L., Sun, G.: Squeeze-and-excitation networks. In: Proceedings of
  the IEEE conference on computer vision and pattern recognition. pp.
  7132--7141 (2018)

\bibitem{huang2018multi}
Huang, G., Chen, D., Li, T., Wu, F., van~der Maaten, L., Weinberger, K.:
  Multi-scale dense networks for resource efficient image classification. In:
  International Conference on Learning Representations (2018)

\bibitem{jia2016dynamic}
Jia, X., De~Brabandere, B., Tuytelaars, T., Gool, L.V.: Dynamic filter
  networks. Advances in neural information processing systems  \textbf{29},
  667--675 (2016)

\bibitem{kang2019few}
Kang, B., Liu, Z., Wang, X., Yu, F., Feng, J., Darrell, T.: Few-shot object
  detection via feature reweighting. In: Proceedings of the IEEE/CVF
  International Conference on Computer Vision. pp. 8420--8429 (2019)

\bibitem{koch2015siamese}
Koch, G., Zemel, R., Salakhutdinov, R., et~al.: Siamese neural networks for
  one-shot image recognition. In: ICML deep learning workshop. vol.~2. Lille
  (2015)

\bibitem{krizhevsky2009learning}
Krizhevsky, A., Hinton, G., et~al.: Learning multiple layers of features from
  tiny images  (2009)

\bibitem{lee2019meta}
Lee, K., Maji, S., Ravichandran, A., Soatto, S.: Meta-learning with
  differentiable convex optimization. In: Proceedings of the IEEE Conference on
  Computer Vision and Pattern Recognition. pp. 10657--10665 (2019)

\bibitem{lee2018gradient}
Lee, Y., Choi, S.: Gradient-based meta-learning with learned layerwise metric
  and subspace. In: International Conference on Machine Learning. pp.
  2927--2936. PMLR (2018)

\bibitem{li2019finding}
Li, H., Eigen, D., Dodge, S., Zeiler, M., Wang, X.: Finding task-relevant
  features for few-shot learning by category traversal. In: Proceedings of the
  IEEE Conference on Computer Vision and Pattern Recognition. pp. 1--10 (2019)

\bibitem{liu2020negative}
Liu, B., Cao, Y., Lin, Y., Li, Q., Zhang, Z., Long, M., Hu, H.: Negative margin
  matters: Understanding margin in few-shot classification. arXiv preprint
  arXiv:2003.12060  (2020)

\bibitem{liu2021learning}
Liu, C., Fu, Y., Xu, C., Yang, S., Li, J., Wang, C., Zhang, L.: Learning a
  few-shot embedding model with contrastive learning. In: Proceedings of the
  AAAI Conference on Artificial Intelligence. vol.~35, pp. 8635--8643 (2021)

\bibitem{liu2018learning}
Liu, Y., Lee, J., Park, M., Kim, S., Yang, E., Hwang, S.J., Yang, Y.: Learning
  to propagate labels: Transductive propagation network for few-shot learning.
  arXiv preprint arXiv:1805.10002  (2018)

\bibitem{liu2020ensemble}
Liu, Y., Schiele, B., Sun, Q.: An ensemble of epoch-wise empirical bayes for
  few-shot learning. In: European Conference on Computer Vision. pp. 404--421.
  Springer (2020)

\bibitem{lu2021tailoring}
Lu, S., Ye, H.J., Zhan, D.C.: Tailoring embedding function to heterogeneous
  few-shot tasks by global and local feature adaptors. In: Proceedings of the
  AAAI Conference on Artificial Intelligence. vol.~35, pp. 8776--8783 (2021)

\bibitem{ma2022adaptive}
Ma, R., Fang, P., Drummond, T., Harandi, M.: Adaptive poincar{\'e} point to set
  distance for few-shot classification. In: Proceedings of the AAAI Conference
  on Artificial Intelligence. vol.~36, pp. 1926--1934 (2022)

\bibitem{nichol2018first}
Nichol, A., Achiam, J., Schulman, J.: On first-order meta-learning algorithms.
  arXiv preprint arXiv:1803.02999  (2018)

\bibitem{oreshkin2018tadam}
Oreshkin, B.N., Rodriguez, P., Lacoste, A.: Tadam: task dependent adaptive
  metric for improved few-shot learning. In: Proceedings of the 32nd
  International Conference on Neural Information Processing Systems. pp.
  719--729 (2018)

\bibitem{qin2020fcanet}
Qin, Z., Zhang, P., Wu, F., Li, X.: Fcanet: Frequency channel attention
  networks. arXiv preprint arXiv:2012.11879  (2020)

\bibitem{Ravi2017OptimizationAA}
Ravi, S., Larochelle, H.: Optimization as a model for few-shot learning. In:
  ICLR (2017)

\bibitem{ren2018meta}
Ren, M., Triantafillou, E., Ravi, S., Snell, J., Swersky, K., Tenenbaum, J.B.,
  Larochelle, H., Zemel, R.S.: Meta-learning for semi-supervised few-shot
  classification. arXiv preprint arXiv:1803.00676  (2018)

\bibitem{ren2015faster}
Ren, S., He, K., Girshick, R., Sun, J.: Faster r-cnn: Towards real-time object
  detection with region proposal networks. Advances in neural information
  processing systems  \textbf{28},  91--99 (2015)

\bibitem{rusu2018meta}
Rusu, A.A., Rao, D., Sygnowski, J., Vinyals, O., Pascanu, R., Osindero, S.,
  Hadsell, R.: Meta-learning with latent embedding optimization. arXiv preprint
  arXiv:1807.05960  (2018)

\bibitem{satorras2018few}
Satorras, V.G., Estrach, J.B.: Few-shot learning with graph neural networks.
  In: International Conference on Learning Representations (2018)

\bibitem{shen2021partial}
Shen, Z., Liu, Z., Qin, J., Savvides, M., Cheng, K.T.: Partial is better than
  all: Revisiting fine-tuning strategy for few-shot learning. In: Proceedings
  of the AAAI Conference on Artificial Intelligence. vol.~35, pp. 9594--9602
  (2021)

\bibitem{shyam2017attentive}
Shyam, P., Gupta, S., Dukkipati, A.: Attentive recurrent comparators. In:
  International Conference on Machine Learning. pp. 3173--3181. PMLR (2017)

\bibitem{simon2020adaptive}
Simon, C., Koniusz, P., Nock, R., Harandi, M.: Adaptive subspaces for few-shot
  learning. In: Proceedings of the IEEE/CVF Conference on Computer Vision and
  Pattern Recognition. pp. 4136--4145 (2020)

\bibitem{simon2020modulating}
Simon, C., Koniusz, P., Nock, R., Harandi, M.: On modulating the gradient for
  meta-learning. In: European Conference on Computer Vision. pp. 556--572.
  Springer (2020)

\bibitem{snell2017prototypical}
Snell, J., Swersky, K., Zemel, R.: Prototypical networks for few-shot learning.
  In: Advances in neural information processing systems. pp. 4077--4087 (2017)

\bibitem{sung2018learning}
Sung, F., Yang, Y., Zhang, L., Xiang, T., Torr, P.H., Hospedales, T.M.:
  Learning to compare: Relation network for few-shot learning. In: Proceedings
  of the IEEE Conference on Computer Vision and Pattern Recognition. pp.
  1199--1208 (2018)

\bibitem{teerapittayanon2016branchynet}
Teerapittayanon, S., McDanel, B., Kung, H.T.: Branchynet: Fast inference via
  early exiting from deep neural networks. In: 2016 23rd International
  Conference on Pattern Recognition (ICPR). pp. 2464--2469. IEEE (2016)

\bibitem{triantafillou2017few}
Triantafillou, E., Zemel, R., Urtasun, R.: Few-shot learning through an
  information retrieval lens. In: Proceedings of the 31st International
  Conference on Neural Information Processing Systems. pp. 2252--2262 (2017)

\bibitem{veit2018convolutional}
Veit, A., Belongie, S.: Convolutional networks with adaptive inference graphs.
  In: Proceedings of the European Conference on Computer Vision (ECCV). pp.
  3--18 (2018)

\bibitem{vinyals2016matching}
Vinyals, O., Blundell, C., Lillicrap, T., Wierstra, D., et~al.: Matching
  networks for one shot learning. In: Advances in neural information processing
  systems. pp. 3630--3638 (2016)

\bibitem{wah2011caltech}
Wah, C., Branson, S., Welinder, P., Perona, P., Belongie, S.: The caltech-ucsd
  birds-200-2011 dataset  (2011)

\bibitem{wang2018skipnet}
Wang, X., Yu, F., Dou, Z.Y., Darrell, T., Gonzalez, J.E.: Skipnet: Learning
  dynamic routing in convolutional networks. In: Proceedings of the European
  Conference on Computer Vision (ECCV). pp. 409--424 (2018)

\bibitem{wang2019simpleshot}
Wang, Y., Chao, W.L., Weinberger, K.Q., van~der Maaten, L.: Simpleshot:
  Revisiting nearest-neighbor classification for few-shot learning. arXiv
  preprint arXiv:1911.04623  (2019)

\bibitem{wang2020generalizing}
Wang, Y., Yao, Q., Kwok, J.T., Ni, L.M.: Generalizing from a few examples: A
  survey on few-shot learning. ACM Computing Surveys (CSUR)  \textbf{53}(3),
  1--34 (2020)

\bibitem{wertheimer2021few}
Wertheimer, D., Tang, L., Hariharan, B.: Few-shot classification with feature
  map reconstruction networks. In: Proceedings of the IEEE/CVF Conference on
  Computer Vision and Pattern Recognition. pp. 8012--8021 (2021)

\bibitem{xu2021learning}
Xu, C., Fu, Y., Liu, C., Wang, C., Li, J., Huang, F., Zhang, L., Xue, X.:
  Learning dynamic alignment via meta-filter for few-shot learning. In:
  Proceedings of the IEEE/CVF Conference on Computer Vision and Pattern
  Recognition. pp. 5182--5191 (2021)

\bibitem{xu2020attentional}
Xu, W., Wang, H., Tu, Z., et~al.: Attentional constellation nets for few-shot
  learning. In: International Conference on Learning Representations (2020)

\bibitem{yan2019meta}
Yan, X., Chen, Z., Xu, A., Wang, X., Liang, X., Lin, L.: Meta r-cnn: Towards
  general solver for instance-level low-shot learning. In: Proceedings of the
  IEEE/CVF International Conference on Computer Vision. pp. 9577--9586 (2019)

\bibitem{ye2020few}
Ye, H.J., Hu, H., Zhan, D.C., Sha, F.: Few-shot learning via embedding
  adaptation with set-to-set functions. In: Proceedings of the IEEE/CVF
  Conference on Computer Vision and Pattern Recognition. pp. 8808--8817 (2020)

\bibitem{zhang2020deepemd}
Zhang, C., Cai, Y., Lin, G., Shen, C.: Deepemd: Few-shot image classification
  with differentiable earth mover's distance and structured classifiers. In:
  Proceedings of the IEEE/CVF Conference on Computer Vision and Pattern
  Recognition. pp. 12203--12213 (2020)

\bibitem{zhao2021looking}
Zhao, J., Yang, Y., Lin, X., Yang, J., He, L.: Looking wider for better
  adaptive representation in few-shot learning. In: Proceedings of the AAAI
  Conference on Artificial Intelligence. vol.~35, pp. 10981--10989 (2021)

\bibitem{zhou2021decoupled}
Zhou, J., Jampani, V., Pi, Z., Liu, Q., Yang, M.H.: Decoupled dynamic filter
  networks. In: Proceedings of the IEEE/CVF Conference on Computer Vision and
  Pattern Recognition. pp. 6647--6656 (2021)

\bibitem{9533455}
Zhu, T., Ma, R., Harandi, M., Drummond, T.: Learning online for unified
  segmentation and tracking models. In: 2021 International Joint Conference on
  Neural Networks (IJCNN). pp.~1--8 (2021)

\bibitem{ziko2020laplacian}
Ziko, I., Dolz, J., Granger, E., Ayed, I.B.: Laplacian regularized few-shot
  learning. In: International Conference on Machine Learning. pp. 11660--11670.
  PMLR (2020)

\end{thebibliography}
